\documentclass[11pt,a4paper]{article}
\usepackage[hyperref]{acl2021}
\usepackage{times}
\usepackage{latexsym}

\usepackage{microtype}

\aclfinalcopy 

\usepackage{amsmath,amsfonts,bm}

\newcommand{\boldtheta}{\boldsymbol{\theta}}
\newcommand{\bolddelta}{\boldsymbol{\delta}}

\def\eqref#1{equation~\ref{#1}}

\def\1{\bm{1}}

\DeclareMathAlphabet{\mathsfit}{\encodingdefault}{\sfdefault}{m}{sl}
\SetMathAlphabet{\mathsfit}{bold}{\encodingdefault}{\sfdefault}{bx}{n}

\usepackage{subcaption}
\usepackage{hyperref}
\usepackage{url}
\usepackage{bbm}
\usepackage[font=small,labelfont=bf]{caption}
\usepackage{booktabs}
\usepackage{graphicx}

\newcommand{\task}{\tau}
\newcommand{\boldz}{\mathbf{z}}
\newcommand{\boldalpha}{\boldsymbol{\alpha}}
\newcommand{\boldw}{\mathbf{w}}
\newcommand{\bolds}{\mathbf{s}}
\newcommand{\boldu}{\mathbf{u}}
\newcommand{\boldone}{\mathbf{1}}
\newcommand{\boldzero}{\mathbf{0}}
\usepackage{latexsym}
\usepackage{multirow}
\newcommand\blfootnote[1]{%
  \begingroup
  \renewcommand\thefootnote{}\footnote{#1}%
  \addtocounter{footnote}{-1}%
  \endgroup
}
\newcommand{\rra}[1]{\renewcommand{\arraystretch}{#1}}

\usepackage{amsmath}
\title{Parameter-Efficient Transfer Learning with Diff Pruning}

\author{Demi Guo \\
  Harvard University   \\
  \textsf{\fontsize{10.5}{10}\selectfont dguo@college.harvard.edu} \\\And 
  Alexander M. Rush\\
  Cornell University \\
  \textsf{\fontsize{10.5}{10}\selectfont  arush@cornell.edu}
  \And
  Yoon Kim \\
  MIT CSAIL \\
  MIT-IBM Watson AI \\
  \textsf{\fontsize{10.5}{10}\selectfont  yoonkim@mit.edu} \\  
  }

\date{}

\begin{document}
\maketitle
\begin{abstract}

The large size of pretrained networks makes them difficult to deploy for multiple tasks in storage-constrained settings.
\emph{Diff pruning} enables parameter-efficient transfer learning that scales well with new tasks. The approach learns a task-specific ``diff" vector that extends the original pretrained parameters. This diff vector is adaptively pruned during training with a differentiable approximation to the $L_0$-norm penalty to encourage sparsity. As the number of tasks increases, diff pruning remains parameter-efficient, as it requires storing only a small diff vector for each task. Since it does not require access to all tasks during training, it is attractive in on-device deployment settings where tasks arrive in stream or even from different providers. Diff pruning can match the performance of finetuned baselines on the GLUE benchmark while only modifying 0.5$\%$ of the pretrained model's parameters per task and scales favorably in comparison to popular pruning approaches.
\vspace{-2mm}
\blfootnote{\noindent \hspace{-6mm} Code:
\textsf{\fontsize{8.5}{10}\selectfont \href{https://github.com/dguo98/DiffPruning}{https://github.com/dguo98/DiffPruning}}}
\end{abstract}

\section{Introduction}
\vspace{-1mm}
Task-specific finetuning of pretrained deep networks is the dominant paradigm
in contemporary NLP, achieving state-of-the-art results across a suite of natural language understanding tasks \citep{devlin2018bert,liu2019roberta,yang2019xlnet,lan2020albert}. While straightforward and empirically effective, this approach is difficult to scale to multi-task, memory-constrained settings (e.g. for on-device applications), as it requires shipping and storing a full set of model parameters \textit{for each task}. Inasmuch as these models are learning generalizable, task-agnostic language representations through self-supervised pretraining, finetuning the entire model for each task seems especially profligate.

 A popular approach to parameter-efficiency is to learn smaller compressed models for each task~\citep{gordon2020compress,sajjad2020poor,zhao2020masking,sanh2020pruning}. Such approaches face a steep sparsity/performance tradeoff and keep a substantial amount of nonzero parameters per task (e.g. 10\%-30\%). Multi-task learning and feature-based transfer allow for more parameter-efficient transfer learning per task~\citep{liu2019multi,clark2019bam,stickland2019pal,reimers2019sbert}. These methods train a small number of additional parameters (e.g. a linear layer) on top of a shared model. However, multi-task learning generally requires access to all tasks during training to prevent catastrophic forgetting~\citep{french1999forgetting}, while feature-based transfer learning (e.g. based on task-agnostic sentence representations) is typically outperformed by finetuning~\citep{howard2018universal}.
 
 An appealing middle ground is to finetune an \textit{extension} of the base model for specific tasks. This approach captures the training benefits of finetuning while maintaining the task modularity of feature-based transfer. For example, Adapters~\citep{rebuffi2018adap} use smaller, task-specific modules that are inserted between layers of a model This approach does not require access to all tasks during training, targeting realistic settings where as new tasks arrive in stream~\citep{houlsby2019adapters,pfeiffer2020fusion,pfeiffer2020adapterhub,pfeiffer2020madx}. 
 \citet{houlsby2019adapters} find that adapter layers can match the performance of fully finetuned BERT on the GLUE benchmark  while requiring 3.6\% additional parameters (on average) per task.

\emph{Diff pruning} is a new extension to pretrained models with the goal of even more parameter-efficient transfer learning. Instead of modifying the architecture of the model, diff pruning 
extends the base model through a task-specific \underline{diff}erence  vector. 

 In order to learn this vector, we reparameterize the task-specific model parameters as $\boldtheta_\text{task} = \boldtheta_\text{pretrained} + \bolddelta_\text{task}$, where the pretrained parameter vector $\boldtheta_\text{pretrained}$ is fixed  and  the task-specific diff vector $\bolddelta_\text{task}$ is finetuned. The diff vector is regularized with a differentiable approximation to the $L_0$-norm penalty~\citep{louizos2018sparse} to encourage sparsity. 
 
 Diff pruning can become extremely parameter-efficient, as it only requires storing the nonzero positions and weights of the diff vector for each task. The cost of storing the shared pretrained model remains constant and is amortized across multiple tasks.
 On the GLUE benchmark~\citep{wang-etal-2018-glue}, diff pruning can match the performance of the fully finetuned BERT baselines  while finetuning only $0.5\%$ of the pretrained parameters per task. As the number of tasks increase,
 diff pruning outperforms popular pruning-based methods in amount of storage required. 

\section{Background: Transfer Learning}

\emph{Transfer learning} in NLP mostly uses a pretrain-and-finetune paradigm, which initializes a subset of the model parameters for all tasks from a pretrained model and then finetunes on a task-specific objective. Pretraining objectives include context prediction \citep{Mikolov2013a}, autoencoding \citep{Dai2015}, machine translation \citep{NIPS2017_7209},  and more recently, variants of language modeling  \citep{peters2018elmo,radford2018gp,devlin2018bert} objectives.

Here we consider applying transfer learning to multiple tasks. We consider a setting with a potentially unknown set of tasks (which may arrive in stream), where each task $\task \in \mathcal{T}$ has an associated training set $\mathcal{D}_\tau = \{x^{(n)}_\task, y^{(n)}_\task \}_{n=1}^N$. For all tasks, the goal is to produce (possibly tied) model parameters $\boldtheta_\task$ to minimize the empirical risk,
\vspace{-3mm}
\begin{align*}
    \min_{\boldtheta_{\task}} \,\, \frac{1}{N} \sum_{n=1}^N C\left(f_\tau(x_\task^{(n)}; \boldtheta_\task), y_\task^{(n)}\right) + \lambda R( \boldtheta_\task ) \\[-6mm]
\end{align*} 
where $f_\tau(\cdot ; \boldtheta_\task)$ is a parameterized function over the input (e.g. a neural network), $C(\cdot, \cdot)$ is a loss function (e.g. cross-entropy),\footnote{While the loss function can be in principle task-specific, in practice we use cross entropy for all tasks and hence omit the subscript in  $C(\cdot, \cdot)$.} and $R(\cdot)$ is an optional regularizer with hyperparameter $\lambda$. 

We can use the pretrain-finetune approach by simply learning independent parameters for each task. However, the large size of pretrained models makes this approach exceedingly parameter inefficient. For example, widely-adopted models such as BERT$_{\textsc{BASE}}$ and BERT$_{\textsc{LARGE}}$ have 110M and 340M parameters respectively, while their contemporaries have parameter counts in the billions \citep{raffel20205t5,shoeybi2019mega,raj2019turing}. Storing the fully finetuned models therefore becomes difficult even for a moderate number of tasks.\footnote{An intriguing line of work suggests that  large-scale language models can be used \emph{without} finetuning for a variety of tasks if given the appropriate context \citep{radford2019language,brown2020language}. While interesting, these models generally underperform task-specific models and require billions of parameters, though recent work suggests that they can be made substantially smaller \citep{schick2020size}.} 
A classic approach to tackling this parameter-inefficiencyis to train a single shared model (along with a task-specific output layer) against multiple tasks through joint training~\citep{caruana1997mtl}. However, the usual formulation of multi-task learning requires the set of tasks $\mathcal{T}$ to be known in advance in order to prevent catastrophic forgetting \citep{french1999forgetting},\footnote{However, work on \emph{continual learning}  mitigates these issues to an extent \citep{NIPS2017_6892,GradientEpisodicMemory,NIPS2017_7051,kirkpatrick2017pnas}.} making it unsuitable for applications in which the set of tasks is unknown or when tasks arrive in stream. 

\vspace{-1mm}
\section{Diff Pruning}
\vspace{-1mm}
 Diff pruning formulates task-specific finetuning as learning a diff vector $\bolddelta_\task$ that is added to the pretrained model parameters $\boldtheta_\text{}$, which remain fixed. We first reparameterize the task-specific model parameters,
 \vspace{-2mm}
\begin{align*}
    \boldtheta_\task = \boldtheta_\text{} + \bolddelta_\task, \\[-6mm]
\end{align*}
which results in the following empirical risk minimization problem,
 \vspace{-2mm}
\begin{align*}
    \min_{\bolddelta_{\task}}\,\,  L(\mathcal{D}_\tau, f_\tau, \boldtheta_\text{} + \bolddelta_\task) + \lambda R( \boldtheta_\text{} + \bolddelta_\tau ),  \\[-6mm]
\end{align*} 
where for brevity we define $L(\mathcal{D}_\tau, f_\tau, \boldtheta_\task)$ as
 \vspace{-2mm}
\begin{align*}
L(\mathcal{D}_\tau, f_\tau, \boldtheta_\task) =  \frac{1}{N} \sum_{n=1}^N C\left(f_\task(x_\task^{(n)}; \boldtheta_\task), y_\task^{(n)}\right).\\[-6mm]
\end{align*}
This trivial reparameterization shows that the cost of storing the pretrained parameters $\boldtheta_\text{}$ is amortized across tasks, and the only 
marginal cost for new tasks is the diff vector. If we can regularize $\bolddelta_\task$ to be sparse such that $\Vert \bolddelta_\task \Vert_0 \ll \Vert \boldtheta_\text{} \Vert_0 $, then this approach can become more parameter-efficient as the number of tasks increases.
We can specify this goal with an $L_0$-norm penalty on the diff vector, 
\vspace{-3mm}
\begin{align*}
    R( \boldtheta_\text{} + \bolddelta_\tau ) = \Vert \bolddelta_\task \Vert_0 = \sum_{i=1}^d \mathbbm{1}\{\bolddelta_{\tau, i} \ne 0\}.
\end{align*}
\vspace{-6mm}
\subsection{Differentiable approximation to the $L_0$-norm}
\vspace{-1mm}
This regularizer is difficult to optimize as it is non-differentiable. In order to approximate this $L_0$ objective, we follow an approach for gradient-based learning with $L_0$ sparsity using a relaxed mask vector~\citep{louizos2018sparse}. 
This approach involves relaxing a binary vector into continuous space, and then multiplying it with a dense weight vector to determine how much of the weight vector is applied during training. After training, the mask is made deterministic, and a large portion of the diff vector is zero.\footnote{It is also possible to learn sparse diff vectors through other penalties such as the $L_1$-norm. We chose to work with the relaxed $L_0$-norm formulation as past work has shown that SGD-based optimization works well in this setting.}

To apply this method we first decompose $\bolddelta_\task$ into a binary mask vector multiplied with a dense vector, 
\vspace{-3mm}
\begin{align*}
    \bolddelta_\task = \boldz_\task \odot \boldw_\task, &&  \boldz_\task \in \{0,1\}^d, \boldw_\task \in \mathbb{R}^d. \\[-6mm]
\end{align*} We now lower bound the true objective and optimize an expectation with respect to $\boldz_\task$, whose distribution $p(\boldz_\task ; \boldalpha_\task)$ is initially Bernoulli with introduced parameters $\boldalpha_\task$,\vspace{-2mm}
\begin{align*} 
\min_{\boldalpha_\task, \boldw_\task}\,\, \mathbb{E}_{\boldz_\task \sim p(\boldz_\task ; \boldalpha_\task)} 
\big[ L(\mathcal{D}_\tau, f_\tau, \boldtheta_\text{} + \bolddelta_\task) 
    +  \lambda \Vert \bolddelta_\task \Vert_0 \big]. \\[-6mm]
\end{align*} 
This objective is still complicated by the discrete nature of $\boldz_\task$'s, but the expectation provides some guidance for 
empirically effective relaxations. We follow prior work \citep{louizos2018sparse,wang2019structuredpruning} and relax $\boldz_\task$ into continuous space $[0, 1]^d$ with a stretched Hard-Concrete distribution~\citep{Jang2017,maddison2017concrete}, which allows for the use of pathwise gradient estimators. Specifically, $\boldz_\task$ is now defined to be a deterministic and (sub)differentiable function of a sample $\boldu$ from a uniform distribution,\vspace{-2mm}
\begin{align*}
&\boldu \sim U(\boldzero, \boldone),\\
   & \bolds_\task = \sigma\left(\log \boldu - \log (1-\boldu) + \boldalpha_\task\right), \\
   & \bar{\bolds}_\task = \bolds_\task \times (r -l) + l, \\
   &\boldz_\task = \min(\boldone, \max(\boldzero, \bar{\bolds}_\task)).  \\[-6mm]
\end{align*}
Here $l < 0$ and $r > 1$ are two constants used to stretch $\bolds_\tau$ into the interval $(l, r)^d$ before it is clamped to $[0, 1]^d$ with the $\min(\boldone, \max(\boldzero, \cdot))$ operation. In this case we have a differentiable closed-form expression for the expected $L_0$-norm, 
\vspace{-3mm}
\begin{align*}
    \mathbb{E}\left[\Vert \bolddelta_\task \right \Vert_0] =
    \sum_{i=1}^d \sigma\left(\boldalpha_{\tau,i} - \log \frac{-l}{r} \right).\\[-6mm]
\end{align*}
Thus the final optimization problem is given by, 
\begin{align*}
    \min_{\boldalpha_\task, \boldw_\task}\, \mathbb{E}_{\boldu \sim U[\boldzero, \boldone]}\left[  L(\mathcal{D}_\tau, f_\tau, \boldtheta_\text{} + \boldz_\task \odot \boldw_\task) \right] 
    \\ +  \lambda \sum_{i=1}^d \sigma\left(\boldalpha_{\tau,i} - \log \frac{-l}{r} \right) ,
\end{align*} 
and we can now utilize pathwise gradient estimators to optimize the first term with respect to $\boldalpha_\task$ since the expectation no longer depends on it.\footnote{To reduce notation clutter we subsume the parameters of the task-specific output layer, which is not pretrained, into $\boldtheta_\text{}$. We do not apply the $L_0$-norm penalty on these parameters during training.}  
After training we obtain the final diff vector $\bolddelta_\tau$ by sampling $\boldu$ once to obtain $\boldz_\task$ (which is not necessarily a binary vector but has a significant number of dimensions equal to exactly zero due to the clamping function), then setting $\bolddelta_\task = \boldz_\task \odot \boldw_\task$.\footnote{We found sampling once to work as well as other alternatives (e.g. based on multiple samples).} 

\vspace{-1mm}
\subsection{$L_0$-ball projection with magnitude pruning for sparsity control}

Differentiable $L_0$ regularization allows us to achieve a high sparsity rate.  However, it would be ideal to set an exact sparsity rate, especially considering applications which require parameter budgets. As  the regularization coefficient $\lambda$ is a Lagrangian multiplier for the constraint $\mathbb{E}\left[\Vert \bolddelta_\task \right \Vert_0] < \eta $ for some $\eta$, this could be achieved in principle by searching over different values of $\lambda$. However we found it more efficient and empirically effective to achieve an exact sparsity rate by projecting onto a target $L_0$-ball after training. 

Specifically, we use magnitude pruning on the diff vector $\bolddelta_\tau$ and target a sparsity rate $t$\% by only keeping the top $t\% \times d$ values in $\bolddelta_\tau$.\footnote{
\citet{wang2019structuredpruning} show that it also is possible to inject such a constraint softly into the training objective by regularizing the expected model size towards a certain rate. However, since the constraint is soft this approach also makes it difficult to target an exact sparsity rate.} Note that unlike standard magnitude pruning, this is based on the magnitude of the diff vector values and not the model parameters.
We found it important to further finetune $\bolddelta_\tau$ with the nonzero masks fixed to maintain good performance, as is often the case in magnitude pruning~\citep{han2016comp}.
Since this type of parameter-efficiency through projection onto the $L_0$-ball can be applied without adaptive diff pruning,\footnote{Concretely, one can obtain $\boldtheta_\tau$ through usual finetuning, set $\bolddelta_\tau = \boldtheta_\tau - \boldtheta_\text{}$, and then apply magnitude pruning followed by additional finetuning on $\bolddelta_\tau$.} such an approach will serve as one of our baselines in the empirical study.
\vspace{-1mm}
\subsection{Structured Diff Pruning}
\vspace{-1mm}

To allow diff pruning to adapt to the model architecture, 
we consider a structured extension which incorporates dependence
between dimensions. We hypothesize that this approach can allow the model
to learn to modify parameters in local regions, as opposed to treating each parameter independently.

We modify the regularizer to first partition the parameter indices into $G$ groups $\{g(1), \dots, g(G)\}$ where $g(j)$ is a subset of parameter indices governed by group $g(j)$.\footnote{While groups can be defined in various ways, we found that defining groups based on each matrix/bias vector of the pretrained model  was simple and worked well enough.} We then introduce a scalar $\boldz_{\tau}^j$ (with the associated parameter $\boldalpha_{\tau}^{j}$) for each group $g(j)$, and decompose the task-specific parameter for index $i \in g(j)$ as
$
    \bolddelta_{\task, i}^j = \boldz_{\task, i} \cdot \boldz_{\task}^j  \cdot \boldw_{\task, i}.
$
The expected $L_0$-norm is then given by
\vspace{-3mm}
\begin{align*}
    & \mathbb{E}\left[\Vert \bolddelta_\task \right \Vert_0] =  \sum_{j = 1}^G \sum_{i \in g(j)}  \mathbb{E}\left[\mathbbm{1}\{\boldz_{\task,i} \cdot \boldz_{\task}^g  > 0 \} \right] \\
   \hspace*{-1cm} &=  \sum_{j = 1}^G \sum_{i \in g(j)}  \sigma\left(\boldalpha_{\tau,i} - \log \frac{-l}{r} \right) 
 \cdot \sigma\left(\boldalpha_{\tau}^j - \log \frac{-l}{r} \right). \\[-7mm]
\end{align*}
We can train with gradient-based optimization as before. Parameters in a group are
encouraged by the regularizer to be removed jointly.

\section{Experiments}

\label{sec:exp}
\subsection{Model and datasets}
\vspace{-1mm}
For evaluation we use the GLUE benchmark \citep{wang2019structuredpruning}
as well as the SQuAD  extractive question answering dataset \citep{rajpurkar-etal-2016-squad}.
Following Adapters \citep{houlsby2019adapters}, we test our approach on the following subset of the GLUE tasks: Multi-Genre Natural Language Inference ({MNLI}), where the goal is two predict whether the relationship between two sentences is entailment, contradiction, or neutral (we test on both MNLI$_m$ and MNLI$_{mm}$ which respectively tests on matched/mismatched domains); Quora Question Pairs ({QQP}), a classification task to predict whether two question are semantically equivalent;  Question Natural Language Inference ({QNLI}), which must predict whether a sentence is a correct answer to the question; Stanford Sentiment Treebank ({SST-2}), a sentence classification task to predict the sentiment of movie reviews; 
Corpus of Linguistic Acceptability ({CoLA}), where the goal is predict whether a sentence is linguistically acceptable or not; Semantic Textual Similarity Benchmark ({STS-B}), which must predict a similarity rating between two sentences; Microsoft Research Paraphrase Corpus ({MRPC}), where the goal is to predict whether two sentences are semantically equivalent; Recognizing Textual Entailment ({RTE}), which must predict whether a second sentence is entailed by the first.
The benchmark uses Matthew's correlation for CoLA, Spearman for STS-B, F$_1$ score for MRPC/QQP, and accuracy for MNLI/QNLI/SST-2/RTE. 

For the main experiments and analysis, we use the $\text{BERT}_{\textsc{LARGE}}$ model from \citet{devlin2018bert} to compare against the adapter-based approach of  \citet{houlsby2019adapters}. Our implementation is based on the Hugging Face Transformer library \citep{Wolf2019HuggingFacesTS}.

\begin{table*}[t]
\rra{1.0}
\begin{minipage}{\textwidth}
\setlength{\tabcolsep}{3pt}
\begin{center}
\resizebox{1.0\textwidth}{!}{

\begin{tabular}{lcccccccccccccc} \toprule 
&  Total & New params  &  & \multirow{2}{*}{QNLI$^{\ast}$} & \multirow{2}{*}{SST-2} & \multirow{2}{*}{$\text{MNLI}_{m}$} & \multirow{2}{*}{$\text{MNLI}_{mm}$} & \multirow{2}{*}{CoLA} & \multirow{2}{*}{MRPC} & \multirow{2}{*}{STS-B} & \multirow{2}{*}{RTE}  & \multirow{2}{*}{QQP}  & \phantom{c} & \multirow{2}{*}{Avg}\\
& params & per task \\
\midrule
Full finetuning              & 9.00$\times$ &  100\%   &    & 91.1 & 94.9  & 86.7   & 85.9    & 60.5 & 89.3 & 87.6  & 70.1 & 72.1 && 80.9    \\

Adapters (8-256)   & 1.32$\times$    &  3.6\% &    & 90.7 & 94.0  & 84.9   & 85.1    & 59.5 & 89.5 & 86.9  & 71.5 & 71.8 & & 80.4    \\ 
Adapters (64)      & 1.19$\times$    &  2.1\% &    & 91.4 & 94.2  & 85.3   & 84.6    & 56.9 & 89.6 & 87.3  & 68.6 & 71.8 & & 79.8    \\ \midrule
Full finetuning  & 9.00$\times$    & 100\% & & 93.4 & 94.1  & 86.7   & 86.0    & 59.6 & 88.9 & 86.6  & 71.2 & 71.7 && 80.6    \\ 
Last layer & 1.34$\times$ & 3.8\% && 79.8 & 91.6 & 71.4 & 72.9 & 40.2 & 80.1 & 67.3 & 58.6 & 63.3 && 68.2\\
Non-adap. diff pruning          & 1.05$\times$   & 0.5\% && 89.7 & 93.6 & 84.9 & 84.8 & 51.2 & 81.5 & 78.2 & 61.5 & 68.6 && 75.5      \\ 
Diff pruning         & 1.05$\times$   & 0.5\%  &&   92.9 & 93.8 & 85.7 & 85.6 & 60.5 & 87.0 & 83.5 & 68.1 & 70.6 && 79.4   \\  
Diff pruning (struct.)     & 1.05$\times$   & 0.5\% & & 93.3 & 94.1  & 86.4   & 86.0    & 61.1 & 89.7 & 86.0  & 70.6 & 71.1 && 80.6    \\ \bottomrule

\end{tabular}}
\vspace{-3mm}
\caption{GLUE benchmark test server results with $\text{BERT}_\textsc{LARGE}$ models. (Top) Results with Adapter bottleneck layers (brackets indicate the size of bottlenecks), taken from from \citet{houlsby2019adapters}. (Bottom) Results from this work. $^\ast$QNLI results are not directly comparable across the two works as the GLUE benchmark has updated the test set since then. To make our results comparable the average column is calculated without QNLI.} \vspace{-2mm}
\label{tab:results}
\end{center} 
\end{minipage}
\end{table*}

\vspace{-2mm}
\subsection{Baselines}
\vspace{-1mm}
We compare both structured and non-structured variants of diff pruning against the following baselines: \textbf{Full finetuning}, which fully finetunes $\text{BERT}_\textsc{LARGE}$ as usual; \textbf{Last layer finetuning}, which only finetunes the penultimate layer (along with the final output layer)\footnote{\citet{wu2020sim} observe that finetuning later layers generally performs better than finetuning earlier layers}; \textbf{Adapters} from \citet{houlsby2019adapters}, which train task-specific bottleneck layers between each layer of a pretrained model, where parameter-efficiency can be controlled by varying the size of the bottleneck layers; and \textbf{Non-adaptive diff pruning}, which performs diff pruning just based on magnitude pruning (i.e., we obtain $\boldtheta_\tau$ through usual finetuning, set $\bolddelta_\tau = \boldtheta_\tau - \boldtheta_\text{}$, and then apply magnitude pruning followed by additional finetuning on $\bolddelta_\tau$). For diff pruning we set our target sparsity rate to 0.5\% and investigate the effect of different target sparsity rates in section~\ref{sec:sparsity}.

\vspace{-2mm}
\subsection{Implementation details and hyperparameters}
\label{sec:implementation}
\vspace{-1mm}
Diff pruning introduces additional hyperparameters $l,r$ (for stretching the Hard-Concrete distribution) and $\lambda$ (for weighting the approximate $L_0$-norm penalty). We found $l =-1.5, r=1.5, \lambda = 1.25 \times 10^{-7}$ to work well across all tasks. We also initialize
the weight vector $\boldw_\task$ to $\boldzero$, and $\boldalpha_\task$ to  a positive vector (we use $\mathbf{5}$) to encourage $\boldz_\task$ to be close to $\boldone$ at the start of training.\footnote{These values were found via by a light hyperparameter search on the SST-2 validation set.} While we mainly experiment with BERT models to faciliate comparison against existing work, in preliminary experiments we found these hyperparameters to work for finetuning RoBERTa \citep{liu2019roberta} and XLNet \citep{yang2019xlnet} models as well.

For all tasks we initially train for 3 epochs and perform a  hyperparameter search over batch size $\in \{5,8,12,16\}$ and learning rate  $\in \{1 \times 10^{-5}, 2 \times 10^{-5},5 \times 10^{-5}\}$.\footnote{However we found the default settings used for  regular finetuning as suggested in the original BERT paper to work well for most tasks.} Finetuning with the fixed mask after projecting onto the $L_0$-ball with magnitude pruning is done for 3 epochs with a learning rate of $5 \times 10^{-5}$ for all datasets except for  MRPC/STS-B/RTE/SST-2 dataset, where we finetune for 5 epochs. The exact hyperparameters for each task are given in section~\ref{sec:glue-hyper} of the appendix. Grouping for the structured version of diff pruning is based on the matrix/bias vectors (i.e. parameters that belong to the same matrix or bias vector are assumed to be in the same group), which  results in 393 groups.\footnote{This definition of groups is implementation-specific since it depends on how one concatenates the input vector before each affine layer. Our grouping is based on Hugging Face's BERT  implementation at commit \textsf{\fontsize{8}{10}\selectfont \href{https://github.com/huggingface/transformers/blob/656e1386a296d696327a9db37de2ccccc79e2cc7/src/transformers/modeling_bert.py}{656e1386a296d696327a9db37de2ccccc79e2cc7}}. We found this simple definition to work well compared to alternative definitions (e.g. based on individual neurons).}

\vspace{-2mm}
\section{Results}
\vspace{-1mm}
\subsection{Results on GLUE}
\vspace{-1mm}

Our main results on the GLUE benchmark are shown in Table~\ref{tab:results}. Structured diff pruning can match the performance of a fully finetuned $\text{BERT}_{\textsc{LARGE}}$ model 
while only requiring 0.5\% additional parameters per task.
Diff pruning without structured sparsity also performs well, though slightly worse than the structured approach. Non-adaptive diff pruning, which magnitude prunes the diff vector without learning the binary mask $\boldz_\task$, performs significantly worse, indicating the importance of learning the masking vector. 
Compared to Adapters, diff pruning obtains similar performance  while requiring many fewer parameters per task, making it a potential alternative for parameter-efficient transfer learning.\footnote{Comparing storage costs is a bit more challenging as it is implementation-specific. Diff pruning incurs additional storage cost due to storing the nonzero positions of the diff vector. See section~\ref{sec:storage} for storage comparison against Adapters assuming float32 for weights and int32 for positions.} 
\vspace{-1mm}
\subsection{Results on SQuAD}
\vspace{-1mm}
To demonstrate the effectiveness of our approach beyond the GLUE tasks, we additionally experiment on SQuAD \citep{rajpurkar-etal-2016-squad}, an extractive question answering dataset where the model has to select the answer span to a question given a Wikipedia paragraph. To make direct comparisons with \citet{houlsby2019adapters}, we run all experiments on SQuAD v1.1. For diff pruning, we use the same general hyperparameters as our full finetuning baseline (see section~\ref{sec:glue-hyper}).
As shown in Figure~\ref{fig:sparsities} (right), diff pruning is able achieve comparable or better performance with only $1.0\%$ additional parameters. Interestingly, diff pruning measurably improves the upon the full finetuning baseline while modifying fewer parameters, which indicates that diff pruning can have a useful regularization effect on top of parameter-efficiency. 
\vspace{-1mm}
\section{Analysis}
\vspace{-1mm}
\subsection{Varying the target sparsity}
\vspace{-1mm}

\begin{figure*}[t]
    \begin{subfigure}{.5\textwidth}
    \centering 
    \includegraphics[width=0.8\textwidth]{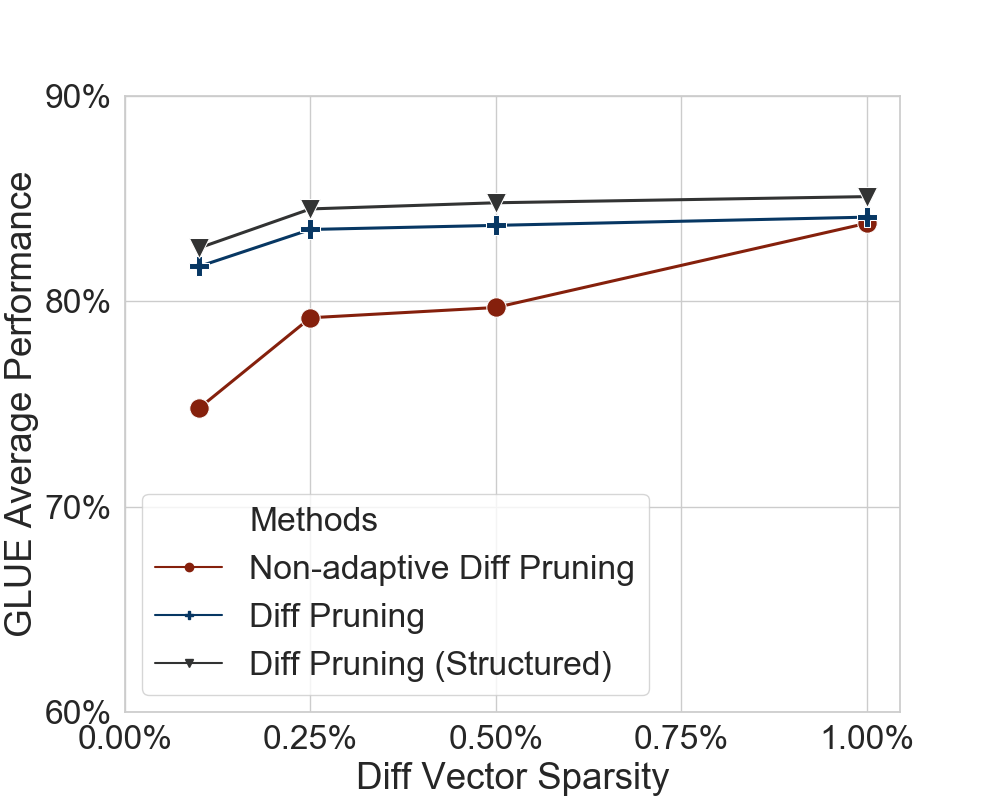}
\end{subfigure}
\begin{subfigure}{.5\textwidth}
\centering
\begin{minipage}{\textwidth}
\begin{center}
\resizebox{0.8\textwidth}{!}{

\begin{tabular}{@{}lcc} \toprule 
& \multicolumn{2}{c}{SQuAD}\\

 & New Params & F$_1$ \\ 
\midrule
\citet{houlsby2019adapters} \\ 
\hspace{3mm} Full finetuning  & 100\% & 90.7 \\
\hspace{3mm} Adapters  & 2.0\%  & 90.4 \\ 
\midrule
This work \\ 
\hspace{3mm} Full finetuning & 100\% & 90.8 \\
\hspace{3mm} Diff pruning  & 1.0\% & 92.1 \\ 
\hspace{3mm} Diff pruning (struct.) & 0.5\% & 91.1 \\ 
\hspace{3mm} Diff pruning (struct.) & 1.0\% & 93.2 \\ 
\bottomrule
\end{tabular}
}
\label{tab:results4}
\end{center} 
\end{minipage}

    \end{subfigure}
    \vspace{-2mm}
    \caption{(Left) Average performance on the GLUE validation set across different target sparsity rates for the different methods. (Right) Results with $\text{BERT}_{\textsc{LARGE}}$ on the SQuAD v1.1 validation set. }
    \vspace{-2mm}
    \label{fig:sparsities}
    \end{figure*}
\label{sec:sparsity}
\begin{table*}[t]
\rra{1.0}

\begin{minipage}{\textwidth}
\begin{center}
\resizebox{1.0\textwidth}{!}{

\footnotesize
\begin{tabular}{@{}ccccccccccccc@{}} \toprule  

Diff vector & &\multirow{2}{*}{QNLI} & \multirow{2}{*}{SST-2} & \multirow{2}{*}{$\text{MNLI}_{m}$} & \multirow{2}{*}{$\text{MNLI}_{mm}$} & \multirow{2}{*}{CoLA} & \multirow{2}{*}{MRPC} & \multirow{2}{*}{STS-B} & \multirow{2}{*}{RTE}  & \multirow{2}{*}{QQP} & & \multirow{2}{*}{Avg}\\
target sparsity  & \\ 
\midrule
0.10\% & &92.7 & 93.3 & 85.6 & 85.9 & 58.0 & 87.4 & 86.3 & 68.6 & 85.2 & & 82.5\\
0.25\% & &93.2 & 94.2 & 86.2 & 86.5 & 63.3 & 90.9 & 88.4 & 71.5 & 86.1 && 84.5 \\
0.50\% & &93.4 & 94.2 & 86.4 & 86.9 & 63.5 & 91.3 & 89.5 & 71.5 & 86.6 && 84.8\\
1.00\% & &93.3 & 94.2 & 86.4 & 87.0 & 66.3 & 91.4 & 89.9 & 71.1 & 86.6 && 85.1 \\ \midrule
100\% &  &93.5 & 94.1  & 86.5   & 87.1   & 62.8 & 91.9 & 89.8 & 71.8 & 87.6  && 85.0\\  

\bottomrule
\end{tabular}
}
\vspace{-2mm}
\caption{Structured diff pruning results on the validation set with different target sparsity rates.}\vspace{-4mm}
\label{tab:results2}

\end{center} 
\end{minipage}
\end{table*}

In Figure~\ref{fig:sparsities} (left), we plot results on the GLUE validation set averaged across all tasks at target sparsity rates of $0.1\%, 0.25\%, 0.5\%, 1.0\%$ for the different baselines. Structured diff pruning consistently outperforms non-structured and and non-adaptive variants across different sparsity rates. The advantage of adaptive methods becomes more pronounced at extreme sparsity rates. In Table~\ref{tab:results2}, we report the breakdown of accuracy of structured diff pruning across different tasks and sparsity rates, where we observe that different tasks have different sensitivity to target sparsity rates. This suggests that we can obtain even greater parameter-efficiency through targeting task-specific sparsity rates in the diff vector.

\vspace{-1mm}
\subsection{Structured vs. Non-structured Diff Pruning}
\vspace{-1mm}
Structured diff pruning introduces an additional mask per group, which encourages pruning of entire groups. This is less restrictive than traditional group sparsity techniques that have been used with $L_0$-norm relaxations, which force all parameters in a group to share the same mask \citep{louizos2018sparse,wang2019structuredpruning}. However we   still expect entire groups to be pruned out more often, which might bias the learning process towards
either eliminating completely or clustering together nonzero diffs. In Table~\ref{tab:results-group}, we indeed find that structured diff pruning leads to finetuned models that are much more likely to leave entire groups unchanged from their pretrained values (zero diffs).

\begin{table*}[t!]
\rra{1.0}

\begin{minipage}{\textwidth}
\begin{center}
\resizebox{0.9\textwidth}{!}{

\footnotesize
\begin{tabular}{@{}cccccccccccc@{}} \toprule  

  & &{QNLI} &{SST-2} & {$\text{MNLI}$} &{CoLA} & {MRPC} & {STS-B} & {RTE}  & {QQP} & & {Avg}  \\
 \midrule 
Non-structured & & 6.2\%  & 6.1\% & 6.0\%  & 6.4\% & 6.1\% & 6.4\% & 7.1\% & 6.1\% & & 6.3\% \\
Structured & & 37.7\% & 64.6\% & 28.8\%  & 20.8\% & 13.2\% & 12.2\% & 12.7\% & 34.9\% && 28.1\%  \\

\bottomrule
\end{tabular}
}
\vspace{-1mm}
\caption{Percentage of groups where all of the parameters in the group are fully zero for structured vs. non-structured diff pruning at 0.5\% target sparsity. We group based on each matrix/bias vector, resulting in 393 groups in total. \vspace{-1mm} }
\label{tab:results-group}
\end{center} 
\end{minipage}
\end{table*}
\vspace{-1mm}
\subsection{Task-specific Sparsity}
\vspace{-1mm}

\begin{figure*}[t]
    \centering
    \includegraphics[width=0.85\textwidth,height=0.27\textwidth   ]{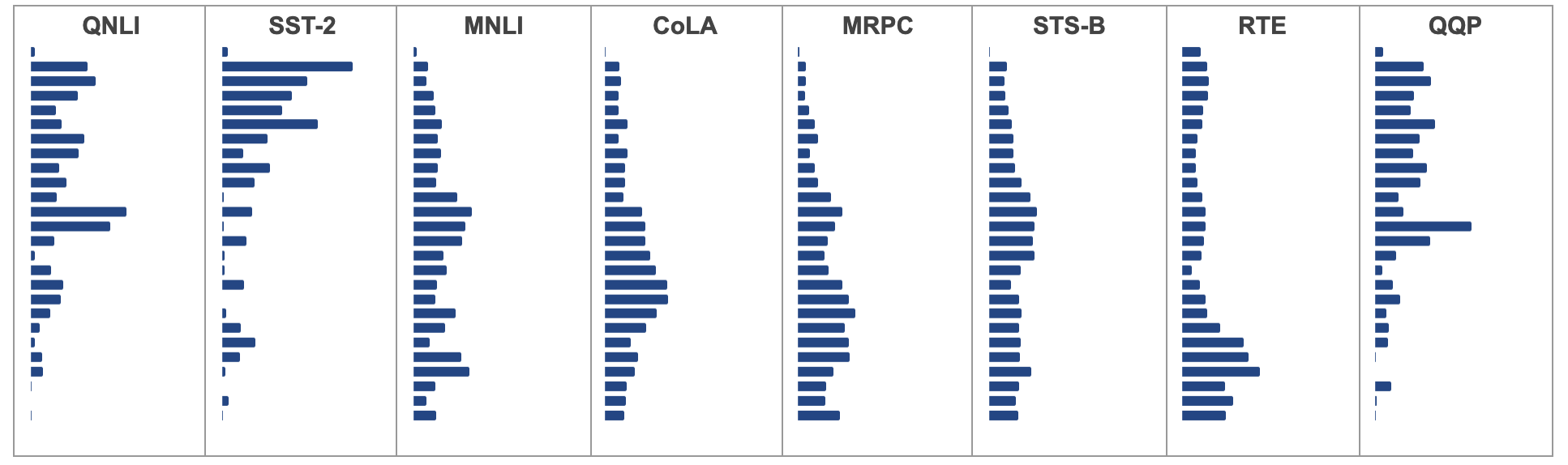}
    \vspace{-2mm}
    \caption{Percentage of modified parameters attributable to each layer for different tasks at 0.5\% target sparsity. The layers are ordered from earlier to later (i.e. the embedding layer is shown at the top). The x-axis for each plot goes from 0\% to 20\%.}
    \vspace{-4mm}
    \label{fig:crosstask}
\end{figure*}
Different layers of  pretrained models have been argued to encode different information \citep{liu_et_al,tenney2019bert}. Given that each task will likely recruit different kinds of language phenomena embedded in the hidden layers, we hypothesize that diff pruning will modify different parts of the pretrained model through task-specific finetuning. Figure~\ref{fig:crosstask} shows the percentage of nonzero diff parameters attributable to the different layers for each task.
We find that different tasks indeed modify different parts of the network, although there are some qualitative similarities between some tasks, for example between QNLI \& QQP (both must encode  questions), and MRPC \& STS-B (both must predict similarity between sentences). The embedding layer is very sparsely modified for all tasks. While some of the variations in the  sparsity distributions is due to simple randomness, we do observe some level of  consistency over multiple runs of the same task, as shown in section~\ref{appendix:random} of the appendix.

The ability to modify different parts of the pretrained model for each task could explain the improved parameter-efficiency of our approach compared to \citet{houlsby2019adapters}'s Adapters, which can only read/write to the pretrained model at certain points of the computational graph.\footnote{To simulate this restricted setting, we tried applying diff pruning only on the  fully-connected layers after the self-attention layers, and observed much worse performance.} This potentially suggests that Adapters with more fine-grained access into model internals (e.g. Adapters for key/value/query transformations) might result in even greater parameter-efficiency. While left as future work, we also note that diff pruning can be applied in conjunction with Adapters, which might further improve results.

\vspace{-1mm}
\subsection{Effect of L$_0$-ball projection}
\vspace{-1mm}
Applying magnitude pruning to project onto the L$_0$-ball was crucial in achieving exact sparsity targets. As shown in Table~\ref{tab:ball-proj}, we observed little loss in performance through this approach. We reiterate that it was crucial to finetune with a fixed mask, even for the approach which does not apply magnitude pruning.\footnote{Without fixed-mask finetuning, GLUE performance decreases from 84.9 to 81.4.}
\begin{table*}[t]
\rra{1.0}

\begin{minipage}{\textwidth}
\begin{center}
\resizebox{1\textwidth}{!}{

\begin{tabular}{@{}lcccccccccccc@{}} \toprule  
 & & QNLI & SST-2 & $\text{MNLI}_{m}$ & $\text{MNLI}_{mm}$ & CoLA & MRPC & STS-B & RTE  & QQP & & Avg\\ 
\midrule

Sparsity w/o Mag. Pruning  & &1.5\% & 0.6\% & 0.8\% & 0.8\% & 1.6\% & 2.4\% & 3.3\% & 0.7\% & 0.6\% && 1.4\%\\
Perf. w/o Mag. Pruning & &93.8 & 94.0 & 86.2 & 86.8 & 63.1 & 91.9 & 89.7 & 71.8 & 86.5 && 84.9\\
\midrule
Mag. Pruning & & 93.4 & 94.2 & 86.4 & 86.9 & 63.5 & 91.3 & 89.5 & 71.5 & 86.6 && 84.8 \\
\bottomrule
\end{tabular}
}
\vspace{-2mm}
\caption{(Top) Sparsity and performance without magnitude pruning on the validation set with structured diff pruning. These results also apply fixed-mask finetuning. (Bottom) Performance with 0.5\% target sparsity and fixed-mask finetuning.}\vspace{-1mm}
\label{tab:ball-proj}

\end{center} 
\end{minipage}

\end{table*}

\vspace{-1mm}
\subsection{Comparison against BERT compression}
\vspace{-1mm}
Direct BERT compression methods also provide a straightforward approach to parameter-efficient transfer learning. Here we compare diff pruning against existing BERT compression methods, in particular DistilBERT \citep{sanh2019distil}, MobileBERT \citep{sun-etal-2020-mobilebert} and TinyBERT \citep{jiao2019tiny}. In these experiments we apply diff pruning on the smaller $\text{BERT}_{\textsc{BASE}}$ model as these works typically utilize $\text{BERT}_{\textsc{BASE}}$ as the baseline.
 As shown in Table~\ref{tab:results-tiny}, we observe that diff pruning is more parameter-efficient when considering all GLUE tasks while maintaining better performance. Of course, BERT compression methods typically have faster inference time (e.g. TinyBERT$_{4}$ is 9.4$\times$ faster that BERT$_{\textsc{BASE}}$). However we note that  diff pruning can be applied on these methods, which may further improve parameter-efficiency while maintaining fast inference.
 

\begin{table*}[t]
\rra{1.0}

\begin{minipage}{\textwidth}
\setlength{\tabcolsep}{3pt}
\begin{center}
\resizebox{1.0\textwidth}{!}{

\begin{tabular}{lcccccccccccccc} \toprule 
&  Total & New params  &  & \multirow{2}{*}{QNLI} & \multirow{2}{*}{SST-2} & \multirow{2}{*}{$\text{MNLI}_{m}$} & \multirow{2}{*}{$\text{MNLI}_{mm}$} & \multirow{2}{*}{CoLA} & \multirow{2}{*}{MRPC} & \multirow{2}{*}{STS-B} & \multirow{2}{*}{RTE}  & \multirow{2}{*}{QQP}  & \phantom{c} & \multirow{2}{*}{Avg}\\
& params & per task \\
\midrule
Full finetuning              & 9.00$\times$ &  100\%   &    & 90.9 & 93.4  & 83.9   & 83.4    & 52.8 & 87.5 & 85.2 & 67.0 & 71.1 && 79.5   \\ 
$\text{DistilBERT}_6$            & 5.53$\times$ &  61.5\%   &    &88.9 & 92.5 & 82.6  & 81.3   & 49.0 & 86.9 & 81.3  & 58.4 & 70.1 && 76.8   \\ 
$\text{TinyBERT}_6$             & 5.53$\times$ &  61.5\%   &    & 90.4 & 93.1  & 84.6   &83.2   & 51.1 & 87.3 & 83.7 & 70.0 & 71.6 && 79.4   \\ 

$\text{DistilBERT}_4$            & 4.31$\times$ &  47.9\%   &    & 85.2 & 91.4 & 78.9  & 78.0  & 32.8 & 82.4 & 76.1 & 54.1 & 68.5 && 71.9   \\ 
$\text{TinyBERT}_4$             & 1.20$\times$ &  13.3\%   &    & 87.7 & 92.6  & 82.5   & 81.8    & 44.1 & 86.4 & 80.4  & 66.6 & 71.3 && 77.0    \\ 

$\text{MobileBERT}_{\text{TINY}}$             & 1.24$\times$ &  13.9\%   &    & 89.5 &91.7 &81.5  & 81.6   & 46.7 & 87.9 & 80.1  & 65.1& 68.9&& 77.0   \\ \midrule

Full finetuning              & 9.00$\times$ &  100\%   &    & 90.9  & 93.4  & 83.9   & 83.5    & 52.1 & 87.9 & 83.6  & 66.2 & 70.7 && 79.1    \\ 
Diff pruning (struct.)     & 1.05$\times$   & 0.5\% & & 90.0 & 92.9  & 83.7   & 83.4    & 52.0 & 88.0 & 84.5  & 66.4 & 70.3 && 79.0    \\ \bottomrule

\end{tabular}}
\vspace{-2mm}
\caption{
Comparison against existing BERT compression works on GLUE. ``Total params" and ``New params per task" columns use $\text{BERT}_{\text{BASE}}$ as the baseline, which has 109M parameters. For example this means that  $\text{MobileBERT}_{\text{TINY}}$ has $13.9\% \times 109\text{M} = 15.1\text{M}$ parameters per task. (Top) Results of different BERT variants, taken from table 1 of \citet{jiao2019tiny}. (Bottom) Structured diff pruning results on $\text{BERT}_{\text{BASE}}$.}\vspace{-4mm}
\label{tab:results-tiny}
\end{center} 
\end{minipage}
\end{table*}

\vspace{-2mm}
\subsection{Storage cost}
\label{sec:storage}
Finally, Table~\ref{tab:analysis-filesize} shows the actual memory requirements for diff pruning compared to Adapters for a Python implementation. While diff pruning requires storing positions in addition to the weights (unlike Adapters which can just store the weights), diff pruning is still more storage-efficient due to the greater parameter-efficiency.
\begin{table}[h!]
\centering
\resizebox{0.5\textwidth}{!}{
\small
\begin{tabular}{l  l l} 
 \toprule
 & New params & Storage (MB) \\
  & per task & per task  \\ 
 \midrule
 Full finetuning  & 100\% &1297.0 \\
 Adapters (weights only) & 3.6\% & 49.0 \\
 Diff pruning  (positions + weights) & 0.5\% & 13.6\\
 \bottomrule
\end{tabular}}
\vspace{-2mm}
\caption{Comparison of file sizes per task based on a basic Python (NumPy) implementation assuming float32 for the weights and int32 for positions.}
\label{tab:analysis-filesize}
\vspace{-3mm}
\end{table}

\vspace{-1mm}
\subsection{Discussion and caveats}
\vspace{-1mm}
 For training, our approach requires more memory than usual finetuning due to additionally optimizing $\boldalpha_\tau$ and $\boldw_\tau$.  Since the majority of GPU memory is typically utilized by a minibatch's intermediate layers, this did not present a significant challenge for pretrained models that we experimented with in this study. However, this could present an issue as model sizes get larger and larger. After training, storing the task-specific diff vector requires storing a compressed version with both the nonzero positions and weights, which incurs additional storage requirements. Finally, while training efficiency was not a primary concern of this work, diff pruning was also approximately $1.5\times$ to $2\times$ slower to train per minibatch than regular finetuning. 

\vspace{-1mm}
\section{Related Work}
\vspace{-1mm}
\paragraph{Multi-task learning} Multi-task learning \citep{caruana1997mtl}, broadly construed, aims to learn models and representations that can be utilized across a diverse range of tasks, and offers a natural approach to training parameter-efficient deep models.  Several works have shown that a single BERT model can obtain good performance  across multiple tasks when jointly trained \citep{liu2019multi,clark2019bam,stickland2019pal}. An alternative approach to multi-task learning that does not require access to all tasks during training involve training smaller task-specific layers that interact with a fixed pretrained model \citep{rebuffi2018adap,Zhang2019sidetuning}. In particular, Adapters \citep{rebuffi2018adap}, which learn to read and write to layers of a shared model, have been applied to obtain parameter-efficient BERT models \citep{houlsby2019adapters,pfeiffer2020fusion,pfeiffer2020adapterhub,pfeiffer2020madx}. In recent work, \citet{li2021prefix} and \citet{qin2021learning} explore the use of learned prompts on top of pretrained models to obtain task-specific models.
Yet another line of work targets extreme parameter-efficiency through task-agnostic sentence representations that can be used without finetuning for downstream tasks \citep{le2014paragraph,kiros2015skipthought,wieting2016sentemb,hill-etal-2016-learning,arora2017simple,conneau2017infersent,cer-etal-2018-universal,zhang-etal-2018-learning-universal,subra2018sent,reimers2019sbert,zhang2020sentence}. These feature-based transfer learning methods are however generally outperformed by fully finetuned models \citep{howard2018universal}.
\vspace{-1mm}
\paragraph{Model compression} There has been much recent work on compressing  pretrained trained with self-supervision (see \cite{ganesh2019compsurvey} for a recent survey). 
A particularly promising line of work focuses on obtaining smaller pretrained models (for subsequent finetuning) through weight pruning \citep{gordon2020compress,sajjad2020poor,chen2020bertlottery} and/or knowledge distillation \citep{sanh2019distil,sun2019patient,turc2019read,jiao2019tiny,sun-etal-2020-mobilebert}. It would be interesting to see whether our approach can be applied on top of these smaller pretrained models to for even greater parameter-efficiency.
\vspace{-1mm}
\paragraph{Learning to mask} Our work is closely related to the line of work on learning to mask parts of deep networks with differentiable relaxations of binary masks for model pruning and parameter sharing \citep{wang2019structuredpruning,zhao2020masking,sanh2020pruning,dixit2020fine,piggyback,Guo_2019_CVPR,sun2020adashare,cao2021low}. While these works also enable parameter-efficient transfer learning, they generally apply the masks directly on the pretrained parameters instead of on the difference vector as in the present work.
\vspace{-2mm}
\paragraph{Regularization towards pretrained models}
Finally, diff pruning is also related to works which regularize the learning process towards pretrained/shared models  for continual learning \citep{rusu106progressive,kirkpatrick2017pnas,schwarz2018progress}, domain adaptation \citep{wiese-etal-2017-neural-domain,miceli-barone-etal-2017-regularization}, and stable finetuning \citep{lee2020mixout}. These works typically do not utilize sparse regularizers and target a different goal than parameter-efficiency.
\vspace{-1mm}
\section{Conclusion}
\vspace{-1mm}
We propose diff pruning as a simple approach for parameter-efficient transfer learning with pretrained models. Experiments on standard NLP benchmarks and models show that diff pruning can match the performance of fully finetuned baselines while requiring only a few additional parameters per task, and can sometimes have a regularization effect and improve upon regular finetuning. We also propose a structured variant of diff pruning which provides further improvements. Avenues for future work include (i) injecting parameter-efficiency objectives directly into the pretraining process (to pretrain models that are better suited towards sparse transfer learning), and (ii) combining diff pruning with other techniques (e.g. adapters, model compression) to achieve even greater parameter-efficiency.

\section*{Acknowledgements}
The authors would like to thank the anonymous reviewers for their valuable feedback on the initial draft. AMR was supported by NSF 1704834 and NSF Career 2037519. 

\bibliography{iclr2021_conference}

\begin{thebibliography}{76}
\providecommand{\natexlab}[1]{#1}
\providecommand{\url}[1]{\texttt{#1}}
\expandafter\ifx\csname urlstyle\endcsname\relax
  \providecommand{\doi}[1]{doi: #1}\else
  \providecommand{\doi}{doi: \begingroup \urlstyle{rm}\Url}\fi

\bibitem[Arora et~al.(2017)Arora, Liang, and Ma]{arora2017simple}
Sanjeev Arora, Yingyu Liang, and Tengyu Ma.
\newblock {A Simple but Tough-to-Beat Baseline for Sentence Embeddings }.
\newblock In \emph{Proceedings of ICLR}, 2017.

\bibitem[Brown et~al.(2020)Brown, Mann, Ryder, Subbiah, Kaplan, Dhariwal,
  Neelakantan, Shyam, Sastry, Askell, Agarwal, Herbert-Voss, Krueger, Henighan,
  Child, Ramesh, Ziegler, Wu, Winter, Hesse, Chen, Sigler, Litwin, Gray, Chess,
  Clark, Berner, McCandlish, Radford, Sutskever, and Amodei]{brown2020language}
Tom~B. Brown, Benjamin Mann, Nick Ryder, Melanie Subbiah, Jared Kaplan,
  Prafulla Dhariwal, Arvind Neelakantan, Pranav Shyam, Girish Sastry, Amanda
  Askell, Sandhini Agarwal, Ariel Herbert-Voss, Gretchen Krueger, Tom Henighan,
  Rewon Child, Aditya Ramesh, Daniel~M. Ziegler, Jeffrey Wu, Clemens Winter,
  Christopher Hesse, Mark Chen, Eric Sigler, Mateusz Litwin, Scott Gray,
  Benjamin Chess, Jack Clark, Christopher Berner, Sam McCandlish, Alec Radford,
  Ilya Sutskever, and Dario Amodei.
\newblock {Language Models are Few-Shot Learners}.
\newblock 2020.

\bibitem[Cao et~al.(2021)Cao, Sanh, and Rush]{cao2021low}
Steven Cao, Victor Sanh, and Alexander~M. Rush.
\newblock {Low-Complexity Probing via Finding Subnetworks}.
\newblock In \emph{Proceedings of NAACL}, 2021.

\bibitem[Caruana(1997)]{caruana1997mtl}
Rich Caruana.
\newblock {Multitask Learning}.
\newblock \emph{Machine Learning}, 1997.

\bibitem[Cer et~al.(2018)Cer, Yang, Kong, Hua, Limtiaco, St.~John, Constant,
  Guajardo-Cespedes, Yuan, Tar, Strope, and Kurzweil]{cer-etal-2018-universal}
Daniel Cer, Yinfei Yang, Sheng-yi Kong, Nan Hua, Nicole Limtiaco, Rhomni
  St.~John, Noah Constant, Mario Guajardo-Cespedes, Steve Yuan, Chris Tar,
  Brian Strope, and Ray Kurzweil.
\newblock Universal sentence encoder for {E}nglish.
\newblock In \emph{Proceedings of EMNLP: System Demonstrations}, 2018.

\bibitem[Chen et~al.(2020)Chen, Frankle, Chang, Liu, Zhang, Wang, and
  Carbin]{chen2020bertlottery}
Tianlong Chen, Jonathan Frankle, Shiyu Chang, Sijia Liu, Yang Zhang, Zhangyang
  Wang, and Michael Carbin.
\newblock {The Lottery Ticket Hypothesis for Pre-trained BERT Networks}.
\newblock \emph{arXiv:2007.12223}, 2020.

\bibitem[Clark et~al.(2019)Clark, Luong, Khandelwal, Manning, and
  Le]{clark2019bam}
Kevin Clark, Minh-Thang Luong, Urvashi Khandelwal, Christopher~D. Manning, and
  Quoc~V. Le.
\newblock {BAM! Born-Again Multi-Task Networks for Natural Language
  Understanding}.
\newblock In \emph{Proceedings of ACL}, 2019.

\bibitem[Conneau et~al.(2017)Conneau, Kiela, Schwenk, Barrault, and
  Bordes]{conneau2017infersent}
Alexis Conneau, Douwe Kiela, Holger Schwenk, Loic Barrault, and Antoine Bordes.
\newblock {Supervised Learning of Universal Sentence Representations from
  Natural Language Inference Data}.
\newblock In \emph{Proceedings of EMNLP}, 2017.

\bibitem[Dai \& Le(2015)Dai and Le]{Dai2015}
Andrew Dai and Quoc~V. Le.
\newblock {S}emi-{S}upervised {S}equence {L}earning.
\newblock In \emph{Proceedings of NIPS}, 2015.

\bibitem[Devlin et~al.(2019)Devlin, Chang, Lee, and Toutanova]{devlin2018bert}
Jacob Devlin, Ming-Wei Chang, Kenton Lee, and Kristina Toutanova.
\newblock {BERT: Pre-training of Deep Bidirectional Transformers for Language
  Understanding}.
\newblock In \emph{Proceedings of NAACL}, 2019.

\bibitem[French(1999)]{french1999forgetting}
Robert French.
\newblock {Catastrophic forgetting in connectionist networks}.
\newblock \emph{Trends in cognitive sciences}, 3, 1999.

\bibitem[Ganesh et~al.(2020)Ganesh, Chen, Lou, Khan, Yang, Chen, Winslett,
  Sajjad, and Nakov]{ganesh2019compsurvey}
Prakhar Ganesh, Yao Chen, Xin Lou, Mohammad~Ali Khan, Yin Yang, Deming Chen,
  Marianne Winslett, Hassan Sajjad, and Preslav Nakov.
\newblock {Compressing Large-Scale Transformer-Based Models: A Case Study on
  BERT}.
\newblock \emph{arXiv:2002.11985}, 2020.

\bibitem[Gordon et~al.(2020)Gordon, Duh, and Andrews]{gordon2020compress}
Mitchell~A. Gordon, Kevin Duh, and Nicholas Andrews.
\newblock {Compressing BERT: Studying the Effects of Weight Pruning on Transfer
  Learning}.
\newblock In \emph{Proceedings of Rep4NLP 2020 Workshop at ACL 2020}, 2020.

\bibitem[Guo et~al.(2019)Guo, Shi, Kumar, Grauman, Rosing, and
  Feris]{Guo_2019_CVPR}
Yunhui Guo, Honghui Shi, Abhishek Kumar, Kristen Grauman, Tajana Rosing, and
  Rogerio Feris.
\newblock {SpotTune: Transfer Learning Through Adaptive Fine-Tuning}.
\newblock In \emph{Proceedings of CVPR}, 2019.

\bibitem[Han et~al.(2016)Han, Mao, and Dally]{han2016comp}
Song Han, Huizi Mao, and William~J. Dally.
\newblock {D}eep {C}ompression: {C}ompressing {D}eep {N}eural {N}etworks with
  {P}runing, {T}rained {Q}uantization and {H}uffman {C}oding.
\newblock In \emph{Proceedings of ICLR}, 2016.

\bibitem[Hill et~al.(2016)Hill, Cho, and Korhonen]{hill-etal-2016-learning}
Felix Hill, Kyunghyun Cho, and Anna Korhonen.
\newblock Learning distributed representations of sentences from unlabelled
  data.
\newblock In \emph{Proceedings of ACL}, 2016.

\bibitem[Houlsby et~al.(2019)Houlsby, Giurgiu, Jastrzebski, Morrone,
  de~Laroussilhe, Gesmundo, and Gelly]{houlsby2019adapters}
Neil Houlsby, Andrei Giurgiu, Stanislaw Jastrzebski, Bruna Morrone, Quentin
  de~Laroussilhe, Andrea Gesmundo, and Mona Attariyanand~Sylvain Gelly.
\newblock Parameter-efficient transfer learning for nlp.
\newblock In \emph{Proceedings of ICML}, 2019.

\bibitem[Howard \& Ruder(2018)Howard and Ruder]{howard2018universal}
Jeremy Howard and Sebastian Ruder.
\newblock {Universal Language Model Fine-tuning for Text Classification}.
\newblock In \emph{Proceedings of ACL}, 2018.

\bibitem[Jang et~al.(2017)Jang, Gu, and Poole]{Jang2017}
Eric Jang, Shixiang Gu, and Ben Poole.
\newblock {C}ategorical {R}eparameterization with {G}umbel-{S}oftmax.
\newblock In \emph{Proceedings of ICLR}, 2017.

\bibitem[Jiao et~al.(2020)Jiao, Yin, Shang, Jiang, Chen, Li, Wang, and
  Liu]{jiao2019tiny}
Xiaoqi Jiao, Yichun Yin, Lifeng Shang, Xin Jiang, Xiao Chen, Linlin Li, Fang
  Wang, and Qun Liu.
\newblock {TinyBERT: Distilling BERT for Natural Language Understanding}.
\newblock In \emph{Proceedings of EMNLP (Findings)}, 2020.

\bibitem[Kirkpatrick et~al.(2017)Kirkpatrick, Pascanu, Rabinowitz, Veness,
  Desjardins, Rusu, Milan, Quan, Ramalho, Grabska-Barwinska, Hassabis, Clopath,
  Kumaran, and Hadsell]{kirkpatrick2017pnas}
James Kirkpatrick, Razvan Pascanu, Neil Rabinowitz, Joel Veness, Guillaume
  Desjardins, Andrei~A. Rusu, Kieran Milan, John Quan, Tiago Ramalho, Agnieszka
  Grabska-Barwinska, Demis Hassabis, Claudia Clopath, Dharshan Kumaran, and
  Raia Hadsell.
\newblock {Overcoming catastrophic forgetting in neural networks}.
\newblock \emph{Proceedings of the National Academy of Sciences}, 14:\penalty0
  3521--3526, 2017.

\bibitem[Kiros et~al.(2015)Kiros, Zhu, Salakhutdinov, Zemel, Torralba, Urtasun,
  and Fidler]{kiros2015skipthought}
Ryan Kiros, Yukun Zhu, Ruslan Salakhutdinov, Richard~S. Zemel, Antonio
  Torralba, Raquel Urtasun, and Sanja Fidler.
\newblock {Skip-Thought Vectors}.
\newblock In \emph{Proceedings of NeurIPS}, 2015.

\bibitem[Lan et~al.(2020)Lan, Chen, Goodman, Gimpel, Sharma, and
  Soricut]{lan2020albert}
Zhenzhong Lan, Mingda Chen, Sebastian Goodman, Kevin Gimpel, Piyush Sharma, and
  Radu Soricut.
\newblock {ALBERT: A Lite BERT for Self-supervised Learning of Language
  Representations}.
\newblock In \emph{Proceedings of ICLR}, 2020.

\bibitem[Le \& Mikolov(2014)Le and Mikolov]{le2014paragraph}
Quoc~V. Le and Tomas Mikolov.
\newblock {Distributed Representations of Sentences and Documents}.
\newblock In \emph{Proceedings of ICML}, 2014.

\bibitem[Lee et~al.(2020)Lee, Cho, and Kang]{lee2020mixout}
Cheolhyoung Lee, Kyunghyun Cho, and Wanmo Kang.
\newblock {Mixout: Effective Regularization to Finetune Large-scale Pretrained
  Language Models}.
\newblock In \emph{Proceedings of ICLR}, 2020.

\bibitem[Lee et~al.(2017)Lee, Kim, Jun, Ha, and Zhang]{NIPS2017_7051}
Sang-Woo Lee, Jin-Hwa Kim, Jaehyun Jun, Jung-Woo Ha, and Byoung-Tak Zhang.
\newblock Overcoming catastrophic forgetting by incremental moment matching.
\newblock In \emph{Advances in Neural Information Processing Systems}. 2017.

\bibitem[Li \& Liang(2021)Li and Liang]{li2021prefix}
Xiang~Lisa Li and Percy Liang.
\newblock {Prefix-Tuning: Optimizing Continuous Prompts for Generation}.
\newblock \emph{arXiv:2101.00190}, 2021.

\bibitem[Liu et~al.(2019{\natexlab{a}})Liu, Gardner, Belinkov, Peters, and
  Smith]{liu_et_al}
Nelson~F. Liu, Matt Gardner, Yonatan Belinkov, Matthew~E. Peters, and Noah~A.
  Smith.
\newblock {Linguistic Knowledge and Transferability of Contextual
  Representations}.
\newblock In \emph{Proceedings of ACL}, 2019{\natexlab{a}}.

\bibitem[Liu et~al.(2019{\natexlab{b}})Liu, He, Chen, and Gao]{liu2019multi}
Xiaodong Liu, Pengcheng He, Weizhu Chen, and Jianfeng Gao.
\newblock {Multi-Task Deep Neural Networks for Natural Language Understanding}.
\newblock In \emph{Proceedings of ACL}, 2019{\natexlab{b}}.

\bibitem[Liu et~al.(2019{\natexlab{c}})Liu, Ott, Goyal, Du, Joshi, Chen, Levy,
  Lewis, Zettlemoyer, and Stoyanov]{liu2019roberta}
Yinhan Liu, Myle Ott, Naman Goyal, Jingfei Du, Mandar Joshi, Danqi Chen, Omer
  Levy, Mike Lewis, Luke Zettlemoyer, and Veselin Stoyanov.
\newblock {RoBERTa: A Robustly Optimized BERT Pretraining Approach}.
\newblock \emph{arXiv:1907.11692}, 2019{\natexlab{c}}.

\bibitem[Lopez-Paz \& Ranzato(2017)Lopez-Paz and
  Ranzato]{GradientEpisodicMemory}
David Lopez-Paz and Marc'Aurelio Ranzato.
\newblock {Gradient Episodic Memory for Continual Learning}.
\newblock In \emph{Proceedings of NeurIPS}, 2017.

\bibitem[Louizos et~al.(2018)Louizos, Welling, P, and
  Kingma]{louizos2018sparse}
Christos Louizos, Max Welling, Diederik P, and Kingma.
\newblock {Learning Sparse Neural Networks through $L_0$ Regularization}.
\newblock In \emph{Proceedings of ICLR}, 2018.

\bibitem[Maddison et~al.(2017)Maddison, Mnih, and Teh]{maddison2017concrete}
Chris~J. Maddison, Andriy Mnih, and Yee~Whye Teh.
\newblock {T}he {C}oncrete {D}istribution: {A} {C}ontinuous {R}elaxation of
  {D}iscrete {R}andom {V}ariables.
\newblock In \emph{Proceedings of ICLR}, 2017.

\bibitem[Mallya et~al.(2018)Mallya, Davis, and Lazebnik]{piggyback}
Arun Mallya, Dillon Davis, and Svetlana Lazebnik.
\newblock {Piggyback: Adapting a Single Network to Multiple Tasks by Learning
  to Mask Weights}.
\newblock In \emph{Proceedings of ECCV}, 2018.

\bibitem[McCann et~al.(2017)McCann, Bradbury, Xiong, and Socher]{NIPS2017_7209}
Bryan McCann, James Bradbury, Caiming Xiong, and Richard Socher.
\newblock Learned in translation: Contextualized word vectors.
\newblock In \emph{Proceedings of NeurIPS}. 2017.

\bibitem[Miceli~Barone et~al.(2017)Miceli~Barone, Haddow, Germann, and
  Sennrich]{miceli-barone-etal-2017-regularization}
Antonio~Valerio Miceli~Barone, Barry Haddow, Ulrich Germann, and Rico Sennrich.
\newblock Regularization techniques for fine-tuning in neural machine
  translation.
\newblock In \emph{Proceedings of EMNLP}, 2017.

\bibitem[Mikolov et~al.(2013)Mikolov, Chen, Corrado, and Dean]{Mikolov2013a}
Tomas Mikolov, Kai Chen, Greg Corrado, and Jeffrey Dean.
\newblock {E}fficient {E}stimation of {W}ord {R}epresentations in {V}ector
  {S}pace.
\newblock \emph{arXiv:1301.3781}, 2013.

\bibitem[Peters et~al.(2018)Peters, Neumann, Iyyer, Gardner, Clark, Lee, and
  Zettlemoyer]{peters2018elmo}
Matthew Peters, Mark Neumann, Mohit Iyyer, Matt Gardner, Christopher Clark,
  Kenton Lee, and Luke Zettlemoyer.
\newblock {D}eep {C}ontextualized {W}ord {R}epresentations.
\newblock In \emph{Proceedings of NAACL}, 2018.

\bibitem[Pfeiffer et~al.(2020{\natexlab{a}})Pfeiffer, Kamath, Ruckle, and amd
  Iryna~Gurevych]{pfeiffer2020fusion}
Jonas Pfeiffer, Aishwarya Kamath, Andreas Ruckle, and Kyunghyun~Cho amd
  Iryna~Gurevych.
\newblock {AdapterFusion: Non-Destructive Task Composition for Transfer
  Learning}.
\newblock \emph{arXiv:2005.00247}, 2020{\natexlab{a}}.

\bibitem[Pfeiffer et~al.(2020{\natexlab{b}})Pfeiffer, Ruckle, Poth, Kamath,
  Vulic, Ruder, and Kyunghyun~Cho]{pfeiffer2020adapterhub}
Jonas Pfeiffer, Andreas Ruckle, Clifton Poth, Aishwarya Kamath, Ivan Vulic,
  Sebastian Ruder, and Iryna~Gurevych Kyunghyun~Cho.
\newblock {AdapterHub: A Framework for Adapting Transformers}.
\newblock \emph{arXiv:2007.07779}, 2020{\natexlab{b}}.

\bibitem[Pfeiffer et~al.(2020{\natexlab{c}})Pfeiffer, Vulic, Gurevych, and
  Ruder]{pfeiffer2020madx}
Jonas Pfeiffer, Ivan Vulic, Iryna Gurevych, and Sebastian Ruder.
\newblock {MAD-X: An Adapter-based Framework for Multi-task Cross-lingual
  Transfer}.
\newblock \emph{arXiv:2005.00052}, 2020{\natexlab{c}}.

\bibitem[Qin \& Eisner(2021)Qin and Eisner]{qin2021learning}
Guanghui Qin and Jason Eisner.
\newblock {Learning How to Ask: Querying LMs with Mixtures of Soft Prompts}.
\newblock In \emph{Proceedings of NAACL}, 2021.

\bibitem[Radford et~al.(2018)Radford, Narasimhan, Salimans, and
  Sutskever]{radford2018gp}
Alec Radford, Karthik Narasimhan, Tim Salimans, and Ilya Sutskever.
\newblock Improving language understanding by generative pre-training.
\newblock 2018.

\bibitem[Radford et~al.(2019)Radford, Wu, Child, Luan, Amodei, and
  Sutskever]{radford2019language}
Alec Radford, Jeff Wu, Rewon Child, David Luan, Dario Amodei, and Ilya
  Sutskever.
\newblock {Language Models are Unsupervised Multitask Learners}.
\newblock 2019.

\bibitem[Radiya-Dixit \& Wang(2020)Radiya-Dixit and Wang]{dixit2020fine}
Evani Radiya-Dixit and Xin Wang.
\newblock {How fine can fine-tuning be? Learning efficient language models}.
\newblock In \emph{Proceedings of AISTATS}, 2020.

\bibitem[Raffel et~al.(2020)Raffel, Shazeer, Adam~Roberts, Narang, Matena,
  Zhou, Li, and Liu]{raffel20205t5}
Colin Raffel, Noam Shazeer, Katherine~Lee Adam~Roberts, Sharan Narang, Michael
  Matena, Yanqi Zhou, Wei Li, and Peter~J. Liu.
\newblock {Exploring the Limits of Transfer Learning with a Unified
  Text-to-Text Transformer}.
\newblock \emph{Journal of Machine Learning Research}, 21, 2020.

\bibitem[Rajbhandari et~al.(2019)Rajbhandari, Rasley, Ruwase, and
  He]{raj2019turing}
Samyam Rajbhandari, Jeff Rasley, Olatunji Ruwase, and Yuxiong He.
\newblock {ZeRO: Memory Optimizations Toward Training Trillion Parameter
  Models}.
\newblock \emph{arXiv:1910.02054}, 2019.

\bibitem[Rajpurkar et~al.(2016)Rajpurkar, Zhang, Lopyrev, and
  Liang]{rajpurkar-etal-2016-squad}
Pranav Rajpurkar, Jian Zhang, Konstantin Lopyrev, and Percy Liang.
\newblock {SQ}u{AD}: 100,000+ questions for machine comprehension of text.
\newblock In \emph{Proceedings of EMNLP}, 2016.

\bibitem[{Rebuffi} et~al.(2018){Rebuffi}, {Vedaldi}, and
  {Bilen}]{rebuffi2018adap}
S.~{Rebuffi}, A.~{Vedaldi}, and H.~{Bilen}.
\newblock {Efficient Parametrization of Multi-domain Deep Neural Networks}.
\newblock In \emph{Proceedings of CVPR}, 2018.

\bibitem[Reimers \& Gurevych(2019)Reimers and Gurevych]{reimers2019sbert}
Nils Reimers and Iryna Gurevych.
\newblock {Sentence-BERT: Sentence Embeddings using Siamese BERT-Networks}.
\newblock In \emph{Proceedings of EMNLP}, 2019.

\bibitem[Rusu et~al.(2016)Rusu, Rabinowitz, Desjardins, Soyer, Kirkpatrick,
  Kavukcuoglu, Pascanu, and Hadsell]{rusu106progressive}
Andrei~A. Rusu, Neil~C. Rabinowitz, Guillaume Desjardins, Hubert Soyer, James
  Kirkpatrick, Koray Kavukcuoglu, Razvan Pascanu, and Raia Hadsell.
\newblock {Progressive Neural Networks}.
\newblock \emph{arXiv:1606.04671}, 2016.

\bibitem[Sajjad et~al.(2020)Sajjad, Dalvi, Durrani, and Nakov]{sajjad2020poor}
Hassan Sajjad, Fahim Dalvi, Nadir Durrani, and Preslav Nakov.
\newblock {Poor Man's BERT: Smaller and Faster Transformer Models}.
\newblock \emph{arXiv:2004.03844}, 2020.

\bibitem[Sanh et~al.(2019)Sanh, Debut, Chaumond, and Wolf]{sanh2019distil}
Victor Sanh, Lysandre Debut, Julien Chaumond, and Thomas Wolf.
\newblock {DistilBERT, a distilled version of BERT: smaller, faster, cheaper
  and lighter}.
\newblock In \emph{Proceedings of 5th Workshop on Energy Efficient Machine
  Learning and Cognitive Computing}, 2019.

\bibitem[Sanh et~al.(2020)Sanh, Wolf, and Rush]{sanh2020pruning}
Victor Sanh, Thomas Wolf, and Alexander~M. Rush.
\newblock {Movement Pruning: Adaptive Sparsity by Fine-Tuning}.
\newblock \emph{arXiv:2005.07683}, 2020.

\bibitem[Schick \& Schutze(2020)Schick and Schutze]{schick2020size}
Timo Schick and Hinrich Schutze.
\newblock {It's Not Just Size That Matters: Small Language Models Are Also
  Few-Shot Learners}.
\newblock \emph{arXiv:2009.07118}, 2020.

\bibitem[Schwarz et~al.(2018)Schwarz, Luketina, Czarnecki, Grabska-Barwinska,
  Teh, Pascanu, and Hadsell]{schwarz2018progress}
Jonathan Schwarz, Jelena Luketina, Wojciech~M. Czarnecki, Agnieszka
  Grabska-Barwinska, Yee~Whye Teh, Razvan Pascanu, and Raia Hadsell.
\newblock {Progress \& Compress: A scalable framework for continual learning}.
\newblock In \emph{Proceedings of ICML}, 2018.

\bibitem[Shin et~al.(2017)Shin, Lee, Kim, and Kim]{NIPS2017_6892}
Hanul Shin, Jung~Kwon Lee, Jaehong Kim, and Jiwon Kim.
\newblock {Continual Learning with Deep Generative Replay}.
\newblock In \emph{Proceedings of NeurIPS}. 2017.

\bibitem[Shoeybi et~al.(2019)Shoeybi, Patwary, Puri, LeGresley, Casper, and
  Catanzaro]{shoeybi2019mega}
Mohammad Shoeybi, Mostofa Patwary, Raul Puri, Patrick LeGresley, Jared Casper,
  and Bryan Catanzaro.
\newblock {Megatron-LM: Training Multi-Billion Parameter Language Models Using
  Model Parallelism}.
\newblock \emph{arXiv:1909.08053}, 2019.

\bibitem[Stickland \& Murray(2019)Stickland and Murray]{stickland2019pal}
Asa~Cooper Stickland and Iain Murray.
\newblock {BERT} and {PAL}s: Projected attention layers for efficient
  adaptation in multi-task learning.
\newblock In \emph{Proceedings of ICML}, 2019.

\bibitem[Subramanian et~al.(2018)Subramanian, Trischler, Bengio, and
  Pal]{subra2018sent}
Sandeep Subramanian, Adam Trischler, Yoshua Bengio, and Christopher~J. Pal.
\newblock {Learning General Purpose Distributed Sentence Representations via
  Large Scale Multi-task Learning}.
\newblock In \emph{Proceedings of ICLR}, 2018.

\bibitem[Sun et~al.(2019)Sun, Cheng, Gan, and Liu]{sun2019patient}
Siqi Sun, Yu~Cheng, Zhe Gan, and Jingjing Liu.
\newblock {Patient Knowledge Distillation for BERT Model Compression}.
\newblock In \emph{Proceedings of EMNLP}, 2019.

\bibitem[Sun et~al.(2020{\natexlab{a}})Sun, Panda, and Feris]{sun2020adashare}
Ximeng Sun, Rameswar Panda, and Rogerio Feris.
\newblock {AdaShare: Learning What To Share For Efficient Deep Multi-Task
  Learning}.
\newblock In \emph{Proceedings of NeurIPS}, 2020{\natexlab{a}}.

\bibitem[Sun et~al.(2020{\natexlab{b}})Sun, Yu, Song, Liu, Yang, and
  Zhou]{sun-etal-2020-mobilebert}
Zhiqing Sun, Hongkun Yu, Xiaodan Song, Renjie Liu, Yiming Yang, and Denny Zhou.
\newblock {M}obile{BERT}: a compact task-agnostic {BERT} for resource-limited
  devices.
\newblock In \emph{Proceedings of ACL}, July 2020{\natexlab{b}}.

\bibitem[Tenney et~al.(2019)Tenney, Das, and Pavlick]{tenney2019bert}
Ian Tenney, Dipanjan Das, and Ellie Pavlick.
\newblock {BERT Rediscovers the Classical NLP Pipeline}.
\newblock In \emph{Proceedings of ACL}, 2019.

\bibitem[Turc et~al.(2019)Turc, Chang, Lee, and Toutanova]{turc2019read}
Iulia Turc, Ming-Wei Chang, Kenton Lee, and Kristina Toutanova.
\newblock {Well-Read Students Learn Better: On the Importance of Pre-training
  Compact Models}.
\newblock \emph{arXiv:1908.08962}, 2019.

\bibitem[Wang et~al.(2019{\natexlab{a}})Wang, Singh, Michael, Hill, Levy, and
  Bowman]{wang-etal-2018-glue}
Alex Wang, Amanpreet Singh, Julian Michael, Felix Hill, Omer Levy, and Samuel
  Bowman.
\newblock {GLUE}: A multi-task benchmark and analysis platform for natural
  language understanding.
\newblock In \emph{Proceedings of ICLR}, 2019{\natexlab{a}}.

\bibitem[Wang et~al.(2019{\natexlab{b}})Wang, Wohlwend, and
  Lei]{wang2019structuredpruning}
Ziheng Wang, Jeremy Wohlwend, and Tao Lei.
\newblock {Structured Pruning of Large Language Models}.
\newblock \emph{arXiv:1910.04732}, 2019{\natexlab{b}}.

\bibitem[Wiese et~al.(2017)Wiese, Weissenborn, and
  Neves]{wiese-etal-2017-neural-domain}
Georg Wiese, Dirk Weissenborn, and Mariana Neves.
\newblock Neural domain adaptation for biomedical question answering.
\newblock In \emph{Proceedings of CoNLL}, August 2017.

\bibitem[Wieting et~al.(2016)Wieting, Bansal, Gimpel, and
  Livescu]{wieting2016sentemb}
John Wieting, Mohit Bansal, Kevin Gimpel, and Karen Livescu.
\newblock {Towards Universal Paraphrastic Sentence Embeddings}.
\newblock In \emph{Proceedings of ICLR}, 2016.

\bibitem[Wolf et~al.(2019)Wolf, Debut, Sanh, Chaumond, Delangue, Moi, Cistac,
  Rault, Louf, Funtowicz, Davison, Shleifer, von Platen, Ma, Jernite, Plu, Xu,
  Scao, Gugger, Drame, Lhoest, and Rush]{Wolf2019HuggingFacesTS}
Thomas Wolf, Lysandre Debut, Victor Sanh, Julien Chaumond, Clement Delangue,
  Anthony Moi, Pierric Cistac, Tim Rault, Rémi Louf, Morgan Funtowicz, Joe
  Davison, Sam Shleifer, Patrick von Platen, Clara Ma, Yacine Jernite, Julien
  Plu, Canwen Xu, Teven~Le Scao, Sylvain Gugger, Mariama Drame, Quentin Lhoest,
  and Alexander~M. Rush.
\newblock Huggingface's transformers: State-of-the-art natural language
  processing.
\newblock \emph{ArXiv}, abs/1910.03771, 2019.

\bibitem[Wu et~al.(2020)Wu, Belinkov, Sajjad, Durrani, Dalvi, and
  Glass]{wu2020sim}
John~M. Wu, Yonatan Belinkov, Hassan Sajjad, Nadir Durrani, Fahim Dalvi, and
  James Glass.
\newblock {Similarity Analysis of Contextual Word Representation Models}.
\newblock In \emph{Proceedings of ACL}, 2020.

\bibitem[Yang et~al.(2019)Yang, Dai, Yang, Carbonell, Salakhutdinov, and
  Le]{yang2019xlnet}
Zhilin Yang, Zihang Dai, Yiming Yang, Jaime Carbonell, Russ~R Salakhutdinov,
  and Quoc~V Le.
\newblock {XLNet: Generalized Autoregressive Pretraining for Language
  Understanding}.
\newblock In \emph{Proceedings of NeurIPS}, 2019.

\bibitem[Zhang et~al.(2020{\natexlab{a}})Zhang, Sax, Zamir, Guibas, and
  Malik]{Zhang2019sidetuning}
Jeffrey~O Zhang, Alexander Sax, Amir Zamir, Leonidas Guibas, and Jitendra
  Malik.
\newblock {Side-Tuning: A Baseline for Network Adaptation via Additive Side
  Networks}.
\newblock In \emph{Proceedings of ECCV}, 2020{\natexlab{a}}.

\bibitem[Zhang et~al.(2018)Zhang, Wu, Li, and
  Li]{zhang-etal-2018-learning-universal}
Minghua Zhang, Yunfang Wu, Weikang Li, and Wei Li.
\newblock Learning universal sentence representations with mean-max attention
  autoencoder.
\newblock In \emph{Proceedings of EMNLP}, 2018.

\bibitem[Zhang et~al.(2020{\natexlab{b}})Zhang, He, Liu, Lim, and
  Bing]{zhang2020sentence}
Yan Zhang, Ruidan He, Zuozhu Liu, Kwan~Hui Lim, and Lidong Bing.
\newblock {An Unsupervised Sentence Embedding Method byMutual Information
  Maximization}.
\newblock In \emph{Proceedings of EMNLP}, 2020{\natexlab{b}}.

\bibitem[Zhao et~al.(2020)Zhao, Lin, Jaggi, and Schutze]{zhao2020masking}
Mengjie Zhao, Tao Lin, Martin Jaggi, and Hinrich Schutze.
\newblock {Masking as an Efficient Alternative to Finetuning for Pretrained
  Language Models}.
\newblock \emph{arXiv:2004.12406}, 2020.

\end{thebibliography}
\bibliographystyle{iclr2021_conference}
\newpage 
\appendix

\begin{table*}[t]
\begin{center}
\resizebox{1.0\textwidth}{!}{

\begin{tabular}{@{}lcccccccccc} \toprule  
 & & QNLI & SST-2 & $\text{MNLI}_{m}$ & $\text{MNLI}_{mm}$ & CoLA & MRPC & STS-B & RTE  & QQP \\ 
\midrule

Learning rate  & & $2 \times 10^{-5}$ & $5 \times 10^{-5}$ & $1 \times 10^{-5}$ & $1 \times 10^{-5}$ & $1 \times 10^{-5}$ & $1 \times 10^{-5}$ & $1 \times 10^{-5}$ & $1 \times 10^{-5}$ & $2 \times 10^{-5}$ \\
Batch size & & 8 & 8 & 8 & 8 & 8 & 8 & 12 & 8 & 8 \\ 
Training epochs & &3  & 3 & 3 & 3 & 3 & 3 & 3 & 3 & 3 \\
Finetuning epochs & & 3 & 5 & 3 & 3 & 3 & 5 & 5 & 5 & 3  \\
\bottomrule
\end{tabular}
}
\vspace{-2mm}
\caption{Best hyperparameters for the GLUE tasks based on the respective validation sets.} \vspace{-10mm}
\label{tab:glue-hyper}
\end{center} 

\end{table*}

\begin{figure*}[t]
    \centering
    \includegraphics[width=0.9\textwidth,height=0.25\textwidth   ]{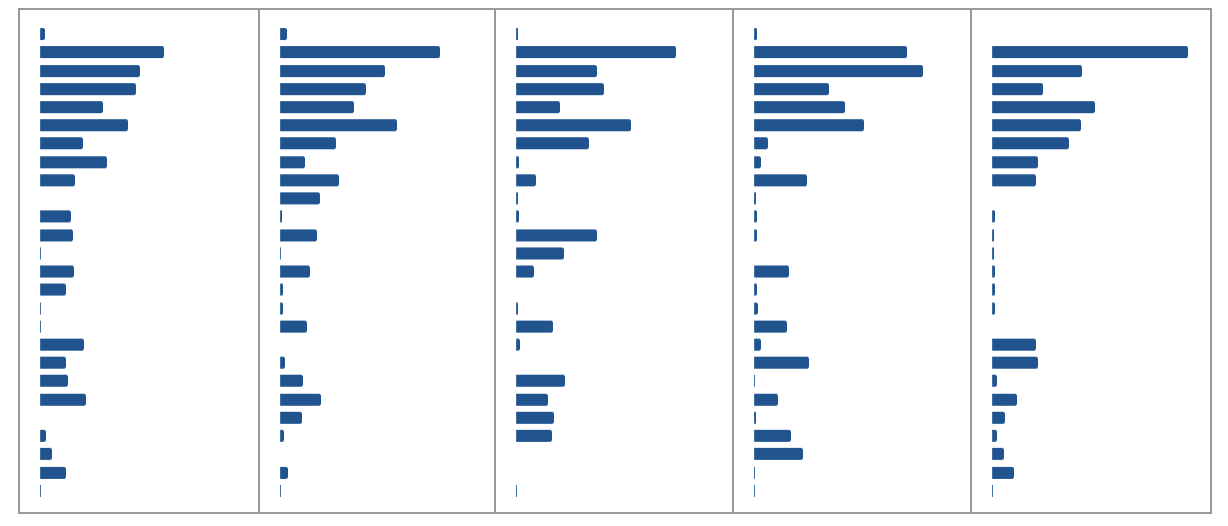}
    \vspace{-3mm}
    \caption{Percentage of modified parameters attributable to each layer for 5 different runs of SST-2 at 0.5\% target sparsity. The layers are ordered from earlier to later (i.e. the embedding layer is shown at the top). The x-axis for each plot goes from 0\% to 20\%.}
    \vspace{-4mm}
    \label{fig:sst-sparsity}
\end{figure*}
\section{Appendix}
\subsection{Hyperparameters}
\label{sec:glue-hyper}
Table \ref{tab:glue-hyper} shows hyperparameters we used for training GLUE tasks. For SQuAD v1.1 experiments, we ran distributed training across 8 GPUs, and used per gpu batch size 3, maximum sequence length 384, document stride 128, learning rate $3 \times 10^{-5}$, number of initial training epochs 2 and number of finetuning epochs 2.

\subsection{Consistency of Nonzero Parameters}
\label{appendix:random}
Figure~\ref{fig:sst-sparsity} shows the percentage of modified parameters attributable to each layer across 5 runs of SST-2. We find that there is nonotrivial variation in sparsity across runs, but also a degree of consistency. For example, the first layer is modified considerably more than other layers across all runs.


\end{document}